\def\year{2022}\relax
\documentclass[letterpaper]{article} 
\usepackage{rgat}  
\pdfoutput=1
\usepackage{times}  
\usepackage{helvet}  
\usepackage{courier}  
\usepackage[hyphens]{url}  
\usepackage{graphicx} 
\usepackage{natbib}  
\usepackage{caption} 
\DeclareCaptionStyle{ruled}{labelfont=normalfont,labelsep=colon,strut=off} 
\usepackage{algorithm}
\usepackage{algorithmic}

\usepackage{subfigure} 
\usepackage{amsmath}
\usepackage{stfloats}
\usepackage{multirow}
\usepackage{enumitem}
\usepackage{booktabs}
\usepackage{bbding}
\usepackage{amsfonts,amssymb}
\usepackage{diagbox}
\usepackage{newfloat}
\usepackage{listings}
\floatstyle{ruled}
\newfloat{listing}{tb}{lst}{}
\floatname{listing}{Listing}
\nocopyright

\title{$\boldsymbol{r}$-GAT: Relational Graph Attention Network for Multi-Relational Graphs}
\author{
    Meiqi Chen \textsuperscript{\rm 1},
    Yuan Zhang \textsuperscript{\rm 1},
    Xiaoyu Kou \textsuperscript{\rm 1},
    Yuntao Li \textsuperscript{\rm 1},
    Yan Zhang \textsuperscript{\rm 1}
}
\affiliations{
    \textsuperscript{\rm 1}Key Laboratory of Machine Perception (MOE), \\
Department of Machine Intelligence, Peking University, Beijing, China\\

    meiqichen@stu.pku.edu.cn\\
    \{yuan.z, kouxiaoyu, liyuntao, zhyzhy001\}@pku.edu.cn
}

\begin{document}

\maketitle

\begin{abstract}
Graph Attention Network (GAT) focuses on modelling \emph{simple} undirected and \emph{single} relational graph data only. This limits its ability to deal with more general and complex multi-relational graphs that contain entities with directed links of different labels (e.g., knowledge graphs). Therefore, directly applying GAT on multi-relational graphs leads to sub-optimal solutions. To tackle this issue, we propose $r$-GAT, a relational graph attention network to learn multi-channel entity representations. Specifically, each channel corresponds to a latent semantic aspect of an entity. This enables us to aggregate neighborhood information for the current aspect using relation features. We further propose a query-aware attention mechanism for subsequent tasks to select useful aspects. Extensive experiments on link prediction and entity classification tasks show that our $r$-GAT can model multi-relational graphs effectively. Also, we show the interpretability of our approach by case study.
\end{abstract}
%

\section{Introduction}
Recent advances of Graph Attention Network (GAT) \cite{velivckovic2017graph} are remarkable in challenging tasks, such as knowledge graph completion \citep{nathani2019learning, zhang2020relational}, text classification \citep{linmei2019heterogeneous}, and dialogue system \citep{qin2020co}, demonstrating its ability to measure the importance of neighbors for graph-structured data.

\begin{figure}
\centering  
\includegraphics[width=0.45\textwidth]{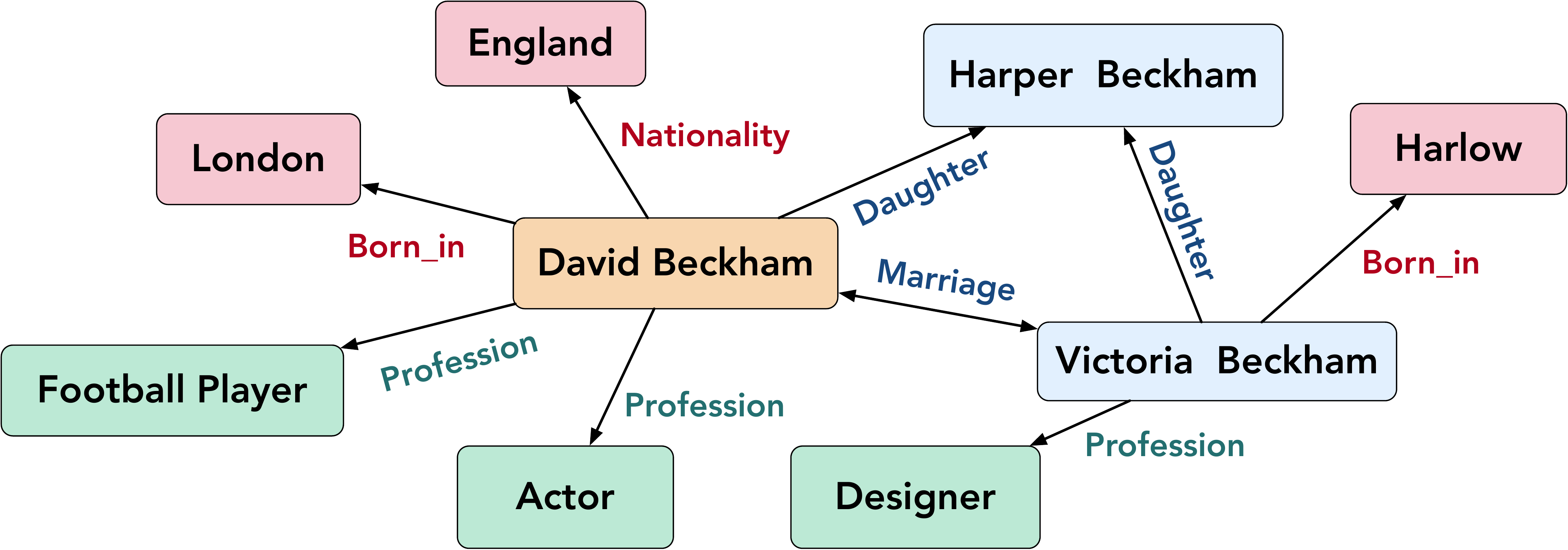}
\caption{An example of a multi-relational graph. \emph{David Beckham}'s neighbors are linked by multiple types of relations.}
\label{fig:example}
\end{figure}

Most of the existing efforts on GAT focus on learning node representation only in graphs with a single relation type (e.g., social networks). In comparison, multi-relational graphs are more general and complex graphs, in which edges are always directed. For example, as shown in Figure \ref{fig:example},  (\emph{David Beckham}, \texttt{born\_in}, \emph{London}) is formulated in triplet form $(s, r, o)$, with $s$ and $o$ representing subject and object entities and $r$ a relation between them. In such graphs, entities and their neighbors are linked by multiple types of relations. These relations usually provide crucial information about the semantic aspects, and the neighbors of \emph{David Beckham} can be classified into three aspects: ``location", ``work", and ``family", according to the semantic information of relations.

Due to various semantic aspects in multi-relational graphs, it is difficult to discriminate the importance of neighbors of an entity using GAT directly. Intuitively, the contribution of a neighbor to an entity may change dynamically, which is affected by different relation types. The relation \texttt{born\_in} and \texttt{nationality} imply the places related to \emph{David Beckham}. Hence, neighbors linked by such two relations would be more important when learning the ``location" aspect of \emph{David Beckham}. In contrast, relation \texttt{profession} is related to the ``work" aspect, so neighbors linked by \texttt{profession} may have higher importance to the ``work" aspect than others. However, GAT ignores the rich semantic information existing in relation features, and learns generic and static entity representations only. As a result, features of different semantic aspects cannot be disentangled for subsequent queries.

In this paper, we propose a novel Relational Graph Attention Network ($r$-GAT) to deal with the multi-relational graphs, which projects entities and relations of a graph into several independent semantic channels. Each channel in $r$-GAT incorporates relation features to measure the importance of neighbors and aggregate neighborhood information related to the corresponding aspect simultaneously. This enables us to disentangle the semantic aspects of entities into multiple components. In order to further leverage the disentangled entity representations, we propose a query-aware attention mechanism to select more relevant aspects for subsequent tasks, so as to get adaptive entity representations. We provide a perspective to show the interpretability of the proposed approach, which is important for downstream applications but not fully explored by previous approaches. 

Our contributions can be summarized as follows:
\begin{itemize}
\item We propose a novel network $r$-GAT, which leverages relation features and learns disentangled entity representations to handle multi-relational graphs.
\item We propose a query-aware attention mechanism to better leverage different semantic aspects for subsequent tasks.
\item Extensive experiments on entity classification and link prediction tasks demonstrate the effectiveness of $r$-GAT, and we show that the proposed approach is more explainable compared with others. 
\end{itemize}

\newcommand{\tabincell}[2]{\begin{tabular}{@{}#1@{}}#2\end{tabular}}
\begin{table*}
\small
    \renewcommand
    \arraystretch{1.0}
    \centering
    \setlength{\tabcolsep}{7pt}
    \begin{tabular}{l|ccccccc}
    \toprule
         \textbf{Methods} & \tabincell{c}{Entity \\ Features} &\tabincell{c}{Weights to \\ Neighbors} & \tabincell{c}{Multi-\\Relations} & \tabincell{c}{Relation \\ Features}& \tabincell{c}{Disentangled \\ Entity Features}  & \tabincell{c}{Query-\\Aware} & \tabincell{c}{Number of Parameters}\\\midrule
         GCN & \Checkmark &\XSolidBrush &\XSolidBrush &\XSolidBrush &\XSolidBrush &\XSolidBrush &$\mathcal{O}(LD_{e}^{2})$\\
         GAT & \Checkmark &\Checkmark &\XSolidBrush &\XSolidBrush &\XSolidBrush &\XSolidBrush &$\mathcal{O}(LD_{e}^{2} + LD_{e})$\\
         R-GCN & \Checkmark &\XSolidBrush &\Checkmark &\XSolidBrush &\XSolidBrush &\XSolidBrush &$\mathcal{O}(\mathcal{B}LD_{e}^{2} + \mathcal{B}L|\mathcal{R}|) $ \\
         COMPGCN & \Checkmark &\XSolidBrush &\Checkmark &\Checkmark &\XSolidBrush &\XSolidBrush &$\mathcal{O}(LD_{e}^{2} + LD_{r}^{2} + \mathcal{B}D_{e} + \mathcal{B}|\mathcal{R}|) $\\
         KBGAT & \Checkmark &\Checkmark &\Checkmark &\Checkmark &\XSolidBrush &\XSolidBrush &$\mathcal{O}(LD_{e}^{2} + LD_{r}^{2} + LD_{e})$\\ 
         \midrule
         $r$-GAT (Ours) & \Checkmark &\Checkmark &\Checkmark &\Checkmark &\Checkmark &\Checkmark & $\mathcal{O}(LD_{e}^{2} + LD_{r}^{2} + LD_{e})$\\
    \bottomrule
    \end{tabular}
    \caption{\label{comp}
    Comparisons between our $r$-GAT and some GNN-based approaches. $L$ is the number of GNN layers, $D_e$ and $D_r$ are the dimensions of entity and relation features. $\mathcal{B}$ denotes the number of bases that R-GCN and COMPGCN use. $|\mathcal{R}|$ denotes the number of relation types. Overall, $r$-GAT is the most expressive approach and is more parameter efficient than the approaches that use relation features.
    }
    \vspace{-4mm}
\end{table*}
\section{Related Work}
\paragraph{Graph Neural Networks}
Recently, Graph Neural Networks (GNNs) are increasingly popular for processing graph-structured data. Compared with the original Graph Convolutional Network (GCN) \citep{kipf2016semi} that treats all neighbors of a node equally, GAT \cite{velivckovic2017graph} measures the importance of neighbors by using an attention mechanism. 

\paragraph{Graph Neural Networks for Multi-Relational Graphs}
RGCN \citep{schlichtkrull2018modeling}, COMPGCN \citep{vashishth2019composition}, and RAGAT \citep{liu2021ragat} transform each neighbor of an entity in terms of the relation between them. When the number of relation types increases, they may easily suffer from over-parameterization. Although they try to alleviate this issue with basis decomposition, the performance is affected by the number of bases. Hence, it is difficult for them to apply to graphs with a large number of relations. Compared with them, $r$-GAT does not introduce dedicated parameters for specific relations, so the problem of over-parameterization is prevented. VR-GCN combines translational properties for multi-relational networks. WGCN \citep{shang2019end} introduces the learnable relation-specific weights to different neighbors. However, it does not consider relation features during the aggregation process. 

KBGAT \citep{nathani2019learning} concatenates entity and relation embeddings in a triplet to calculate the attention values. Then, the new representation of an entity is obtained by summing every weighted triplet representation. Different from KBGAT, $r$-GAT transforms entity and relation features separately. KBGAT uses the entity and relation embeddings produced by TransE \citep{bordes2013translating} to initialize their embeddings, whereas our $r$-GAT learns the entity and relation features from scratch. Moreover, like classic GAT, KBGAT only learns generic representation for every entity and the difference of neighbors’
importance w.r.t. various semantic aspects is ignored. Therefore, it is not adaptive to all the downstream queries. And it is difficult to explain the outcome attention weights of neighbors it obtains. We summarize the difference between our $r$-GAT and other approaches in Table \ref{comp}.

\begin{figure*}
\centering 
\includegraphics[scale=0.3]{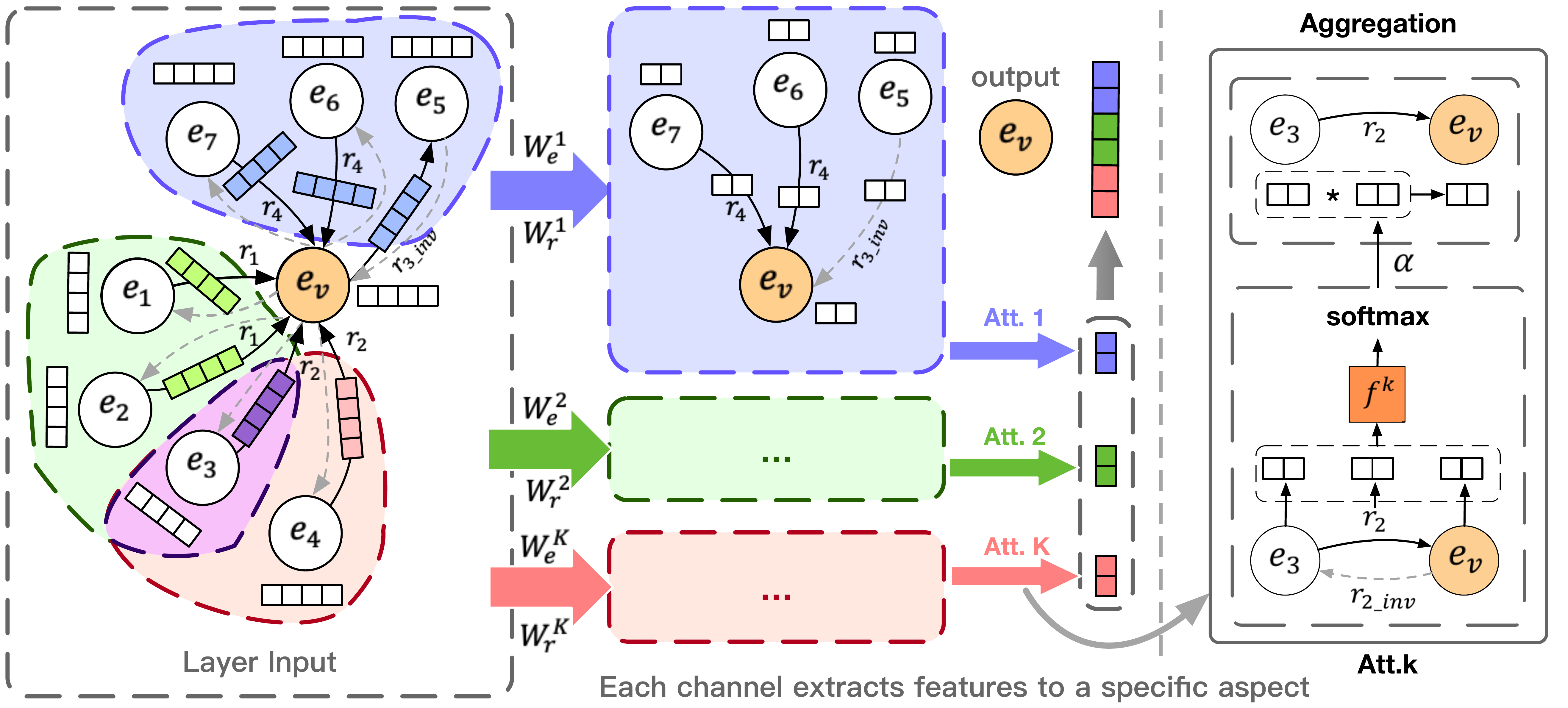} 
\caption{Overview of $r$-GAT (this example assumes that there are $K$ latent semantic aspects). It takes the entity and relation features as the input and outputs disentangled entity representation. $r$-GAT leverages the relation features in the process of neighborhood aggregation. Each channel extracts neighborhood features specific to the current semantic aspect.} 
\label{fig:$r$-GAT} 
\end{figure*}

\paragraph{Knowledge Graph Embedding}
Knowledge Graph Embedding (KGE) has been an active approach for Knowledge Graph Completion (KGC). It embeds entities and relations in a continuous vector or matrix space. The goal is to learn embeddings such that a well-designed scoring function assigns higher scores to valid triplets than the invalid ones. In terms of the different types of scoring function, previous KGE models can be divided into translation models \citep{bordes2013translating, wang2014knowledge}, factorization models \citep{nickel2011three, yang2014embedding, trouillon2016complex, balazevic2019tucker}, and Neural Network-based models \citep{dettmers2018convolutional, xie2020reinceptione, vashishth2020interacte}. A relevant and concurrent work Knowledge Router \citep{zhang2021knowledge} extends DisenGCN \citep{ma2019disentangled}'s neighborhood routing mechanism to knowledge graphs. However, Knowledge Router does not use the relation features in the routing mechanism. And the multiple components they learned are not further leveraged to achieve query-aware for the subsequent tasks.

\section{Approach}
\subsection{Background}
For a multi-relational graph $\mathcal{G} = \{\mathcal{V}, \mathcal{R},  \mathcal{E}\}$ where $\mathcal{V}$, $\mathcal{R}$ represent the set of entities and relations respectively, and $\mathcal{E}$ is the set of edges. Each edge $(s, r, o) \in \mathcal{E}$ denotes entity $s$ has a relation $r$ to entity $o$. Following \citep{schlichtkrull2018modeling}, we add the inverse relations to $\mathcal{R}$ to allow the information flowing in both directions. Hence, $\mathcal{E}$ and $\mathcal{R}$ are extended as:
\begin{equation}
    \mathcal{E}^{\prime}=\mathcal{E} \cup \left\{\left(o, r^{-1}, s\right) \mid (s, r, o) \in \mathcal{E}\right\},
\end{equation}     
and $\mathcal{R}^{\prime}=\mathcal{R} \cup \mathcal{R}_{inv}$, where $\mathcal{R}_{inv} = \left\{ r^{-1} \mid r \in \mathcal{R} \right\}$ denotes the inverse relations. Besides, we add the self-loops to $\mathcal{E}$ and we represent all the relations in self-loops as $r_{\rm sp}$.

\subsection{Relational Graph Attention Network}
$r$-GAT leverages relation features in a multi-relational graph. Because in these graphs, relations provide crucial information about why two entities are connected. Both entity features and relation features are projected into several independent spaces (channels). Each channel identifies the contribution of neighborhood information to the current aspect of an entity. An overview of $r$-GAT is presented in Figure \ref{fig:$r$-GAT}.

We start by describing a single relational graph attention layer, which is the building block of $r$-GAT. The $l$-th layer of $r$-GAT takes a set of entity features $\mathbf{e} \in \mathbb{R}^{N_{e} \times D_{e}^{(l-1)}}$ and relation features $\mathbf{r} \in \mathbb{R}^{N_{r} \times D_{r}^{(l-1)}}$ as input, and outputs a new set of entity and relation features: $\mathbf{e}^{(l)} \in \mathbb{R}^{N_{e} \times D_{e}^{(l)}} $, $\mathbf{r}^{(l)} \in \mathbb{R}^{N_{r} \times D_{e}^{(l)}} $, where $N_e$ and $N_r$ are the numbers of entities and relations, $D_{e}^{(l-1)}$ and $D_r^{(l-1)}$ are the dimensions of input entity and relation features, $D_e^{(l)}$ is the dimensions of output features.

Each layer consists of $K$ channels, we expect that $K$ channels can extract different semantic features by projecting an entity $v$ into $K$ different subspaces. Consider the edge $(v, i, u)$, for each channel, $e_{v}^{k} = \mathbf{W}_{e}^{k}e_{v}$ describes the features of entity $v$ that is relevant to aspect $k$. Then we transform the relation features to the same size as entity's: ${r}_{i}^{k}= \mathbf{W}_{r}^{k}r_{i}$, where $\mathbf{W}_{e}^{k} \in \mathbb{R}^ {\frac{ D_{e}^{(l)}}{K} \times D_{e}}$ and $\mathbf{W}_{r}^{k} \in \mathbb{R}^ {\frac{ D_{e}^{(l)}}{K} \times D_{r}}$ are the parameter weight matrices for the $k$-th channel.

In order to let more semantically related neighbors have higher importance under the current aspect, we consider the relation features and perform a shared attention mechanism when measuring the importance of neighbors for learning the $k$-th component of entity $v$:
\begin{equation}
\label{eq:leaky-relu1}
    {\rm att}_{viu}^{k}=f^{k}\left[e_{v}^{k}||r_{i}^{k} ||e_{u}^{k}  \right],
\end{equation}
where $\|$ represents concatenation, $f^{k}$ is a feedforward neural network that is parameterized by a weight matrix $\mathbf{W}_{f}^{k}\in\mathbb{R}^{1\times \frac{3 \times D_{e}^{(l)}}{K}}$, followed by a nonlinearity. 

To make the importance more comparable, a softmax function is applied over the attention values of channel $k$:
\begin{equation}
\begin{aligned}
    \alpha_{viu}^{k} &=\mathrm{softmax}_{ui}\left({\rm att}_{viu}^{k}\right) \\
    &=\frac{\exp \left({\rm att}_{viu}^{k}\right)}{\sum_{z \in \mathcal{N}_{v}} \sum_{j \in \mathcal{R}_{vz}} \exp \left({\rm att}_{vjz}^{k}\right)},
\end{aligned}
\end{equation}
where $\mathcal{N}_{v}$ are all the 1-order neighbors of entity $v$, $\mathcal{R}_{vz}$ are the relations between entity $v$ and its neighbor $z$. 

$\alpha_{viu}^{k}$ can be seen as contribution of the neighbor $u$ to construct $e_{v}^{k(l)}$. To aggregate information from neighbor $u$ to the $k$-th channel of $v$, we incorporate the relation features $ r_{i}^{k}$ and multiply $e_{u}^{k}$ with it:
\begin{equation}
\label{eq:elu1}
    e_{v}^{k(l)}=\sigma_{1}\left(\sum_{u \in \mathcal{N}_{v}} \sum_{i \in \mathcal{R}_{vu}} \alpha_{viu}^{k} [e_{u}^{k} *  r_{i}^{k}]\right).
\end{equation}
where $\sigma_{1}$ is a nonlinearity, * denotes multiplication.

Finally, the disentangled representation of entity $v$ is obtained by paralleling the calculation of each channel and concatenating all the $K$ channels' features:
\begin{equation}
\label{eq:elu2}
e_{v}^{(l)}=\mathop{\Big{\|}}\limits_{k=1}^{K}\sigma_{1}\left(\sum_{u \in \mathcal{N}_{v}} \sum_{i \in \mathcal{R}_{vu}} \alpha_{viu}^{k} [e_{u}^{k} *  r_{i}^{k}]\right),
\end{equation}
where $\|$ represents concatenation. In this setting, $e_{v}^{(l)}$ consists of $\frac{K \times D_{e}^{(l)}}{K} = D_{e}^{(l)}$ features. By simultaneously computing features for all the entities, a disentangled entity features matrix $\mathbf{e}^{(l)} \in \mathbb{R}^{N_{e}\times D_e^{(l)}}$ is obtained. $r$-GAT is able to extract information of higher-order neighborhood by stacking multiple relational graph attention layers.

\paragraph{Comparison with Classic GAT} Compared with GAT, $r$-GAT takes the relation features into consideration. The motivation of multi-channel in $r$-GAT is different from that of multi-head in GAT. Multi-head attention in GAT is applied mostly to stabilize the learning process. In contrast, $r$-GAT does not employ averaging, it just learns and preserves disentangled entity features. Each channel of $r$-GAT stands for a specific semantic aspect of an entity, which is useful for subsequent tasks. More comparisons between our $r$-GAT and other GNN-based approaches can be found in Table \ref{comp}.

\subsection{Query-Aware for Link Prediction}
Link prediction is the task of predicting missing links based on the known ones, which aims at learning a scoring function and assigning higher scores to the valid triplets than the invalid ones.

For the link prediction task, we use $r$-GAT as the encoder, and various KGE models can be used as the decoder. The initial hidden representations for each entity are set to trainable embedding vectors. $r$-GAT outputs disentangled entity representation matrix $\mathbf{e}^{(L)}$ to the decoder, where $L$ denotes the number of $r$-GAT layers. In this section we abbreviate every entity embedding $e^{(L)}$ as $e$. 

\begin{figure}
\centering 
\includegraphics[width=0.48\textwidth]{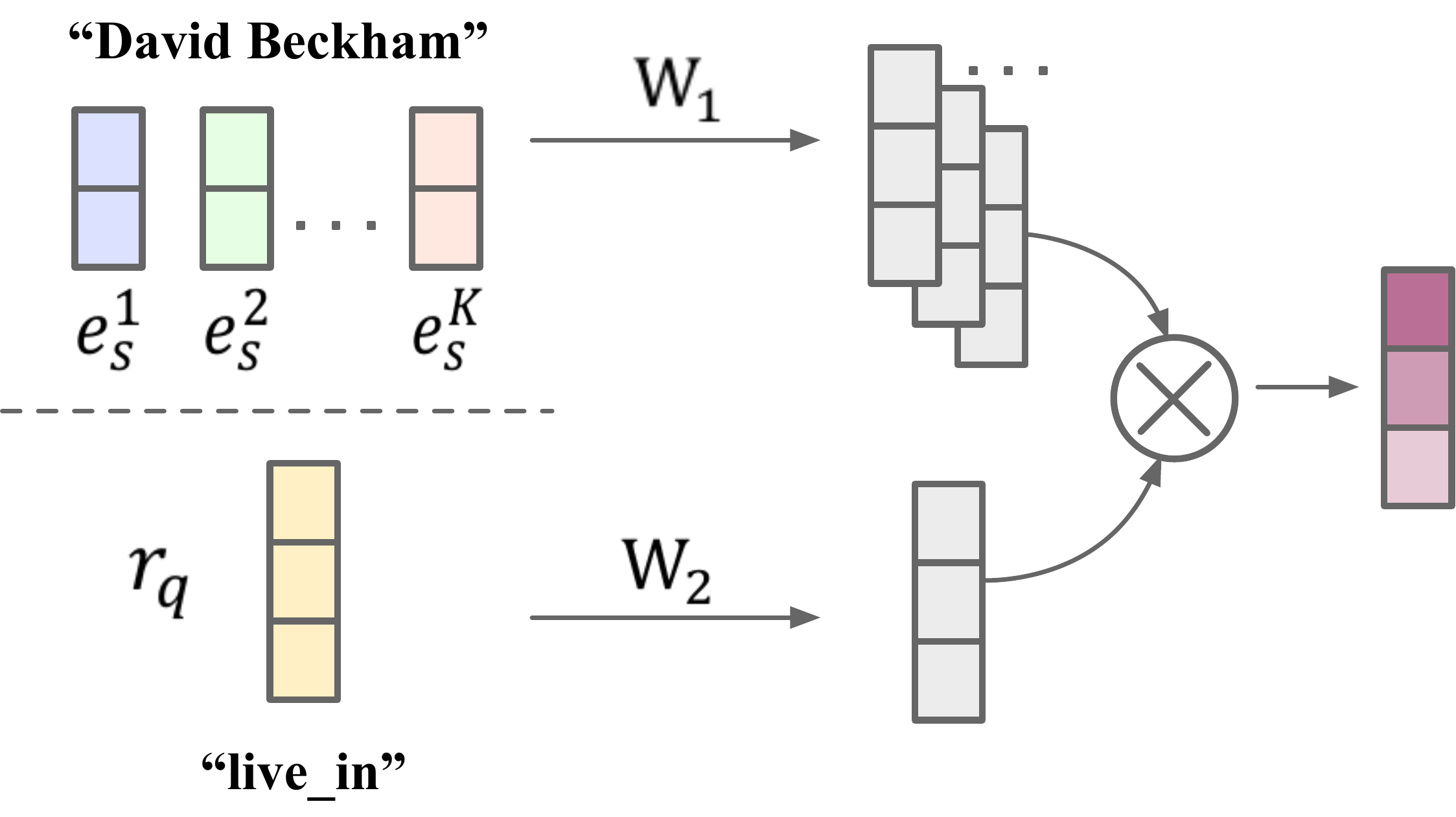} 
\caption{The query-aware attention mechanism. } 
\label{fig:qa} 
\end{figure}

\paragraph{Query-Aware Attention Mechanism (Qatt)} Given a query relation $q$ and a subject entity $s$, as shown in Figure \ref{fig:qa}, before inputting the entity embeddings to the decoder, we first apply an attention mechanism between the query relation and all the $K$ components to capture the more relevant semantic information. 
\begin{equation}
\label{eq:qa}
    {\beta}_{sq}^{k}=\mathrm{softmax}_{k} ( \frac{(\mathbf{W}_{1}e_{s}^{k})^{T} (\mathbf{W}_{2}r_{q}) }{\sqrt{D_{q}}}),
\end{equation}
where ${\beta}_{sq}^{k}$ represents the importance of the $k$-th aspect of entity $s$ to query relation $q$. $\mathbf{W}_{1}\in \mathbb{R}^{D_{q} \times \frac{D_e^{(L)}}{K}}$, $\mathbf{W}_{2}\in \mathbb{R}^{D_{q} \times D_r^{(L)}}$ are parameter weight matrices, $\sqrt{D_{q}}$ is a scaling factor \citep{vaswani2017attention}. 

Then we can obtain the query-aware entity embeddings, which facilitates the disentangling of components and makes our approach adaptive to the various query relations:
\begin{equation}
    \mathcal{Q}(e_{s}, r_q)=\mathop{\Big{\|}}\limits_{k=1}^{K}\left({\beta}_{sq}^{k} (\mathbf{W}_{3}[e_{s}^{k}|| r_{q}])\right),
\end{equation}
where $\mathbf{W}_{3}\in\mathbb{R}^{D_{q} \times (D_r^{(L)} + \frac{D_e^{(L)}}{K})}$, $\|$ represents concatenation. The query-aware attention mechanism can also be extended to a multi-head form to stabilize the learning process. 

Finally, $\mathcal{Q}(e_{s}, r_{q})$ is projected into the entity embedding dimension $D_e^{(L)}$ and matched with the object embedding $e_o$ via a dot product. The scoring function can be written as:
\begin{equation}
\label{eq:relu}
\psi_{r} \left(e_{s}, e_{o}\right)=\sigma_{2}\left(\mathbf{W} {\mathcal{Q}}(e_{s}, r_{q})\right) e_{o},
\end{equation}
where $\mathbf{W}$ is the weight matrix for linear transformation, $\sigma_{2}$ denotes a nonlinearity. 

Following the 1-N scoring procedure proposed by \citet{dettmers2018convolutional}, we score one pair $(e_s, r_q)$ against all the entities $o\in \mathcal{V}$ simultaneously. With a logistic sigmoid to the logits of the scores of $(e_s, r_q, e_o)$, the model is trained by minimizing the Binary Cross-Entropy (BCE) loss:
\begin{equation}
\small
\mathcal{L}=-\frac{1}{N} \sum_{i}\left(t_{i} \cdot \log \left(p_{i}\right)+\left(1-t_{i}\right) \cdot \log \left(1-p_{i}\right)\right),
\end{equation}
where $p_i = \sigma_{3} (\psi_{r} \left(e_{s}, e_{o}\right) )$ is the predicted probability and $t_i$ is the binary label, $\sigma_{3}$ denotes sigmoid.

\begin{table}
    \renewcommand
    \arraystretch{1.0}
    \centering
    \setlength{\tabcolsep}{4pt}
    \begin{tabular}{lcc}
    \toprule
        \textbf{Dataset} & \textbf{FB15k-237} & \textbf{WN18RR} \\ \midrule
       Entities & 14541 & 40943 \\
       Relations &237 &11\\
       Edges &\tabincell{c}{272115 / 17525 / 20466} &\tabincell{c}{86835 / 3034 / 3134} \\
    \bottomrule
    \end{tabular}
    \caption{Statistics of datasets.}
\label{tab:lpdata}
    \vspace{-4mm}
\end{table}
\paragraph{Tackling the Query-Ignorant Problem}Previous GNN-based approaches suffer from the query-ignorant problem: during the neighborhood aggregation process, an entity is unable to know which neighbor is more important without access to future query relations. Hence, those approaches only learn generic entity embeddings, which are not adaptive to all the subsequent query relations. $r$-GAT addresses this by learning disentangled entity embeddings in parallel and in advance. In this way, different semantic aspects are grouped into the corresponding channels, and a query relation can focus more on the relevant aspects of an entity by focusing more on the relevant channels.

\subsection{Entity Classification}
For semi-supervised entity classification task, we stack the layers of $r$-GAT and apply the softmax function on the last layer. Following \citet{schlichtkrull2018modeling}, we minimize the Cross-Entropy (CE) loss on all the labeled entities:
\begin{equation}
\mathcal{L}=-\frac{1}{\mathcal{Y}}\sum_{i} \sum_{c=1}^{C} t_{i c} \ln p_{i c},
\end{equation}
where $\mathcal{Y}$ is the set of labeled entities, $C$ is the number of label types, $p_{i c}$ is the predicted probability, $t_{i c}$ indicates whether entity $i$ belongs to the type $c$.

\section{Experiments}
\label{sec:setup}
In this section, we evaluate our $r$-GAT on two popular tasks for multi-relational graphs: link prediction (predicting missing facts based on the known ones) and entity classification (assigning types or labels to the entities).
\paragraph{Link Prediction Dataset}
We use two benchmark knowledge graph datasets to evaluate the performance of $r$-GAT on link prediction task: \textbf{FB15k-237} \citep{toutanova2015representing} is a subset of FB15k \citep{bordes2013translating}, originally derived from Freebase. FB15k-237 contains the knowledge graph triplets of Freebase entities, but removes the reverse relations, making it more difficult for a model to predict the new facts. \textbf{WN18RR} \citep{dettmers2018convolutional} is created from WN18 \citep{bordes2013translating}, which is a subset of WordNet. \citet{dettmers2018convolutional} find that the test sets of WN18 contain many triplets that can be obtained by simply inverting triplets in the training set, so WN18RR is introduced to avoid the reverse relation test leakage problem. 

In recent years, FB15k-237 and WN18RR have become the most popular datasets for knowledge graph link prediction. The statistics of datasets are shown in Table \ref{tab:lpdata}.
\paragraph{Entity Classification Dataset}
We evaluate $r$-GAT on three standard entity classification datasets: AIFB, MUTAG, and BGS. The details of datasets are shown in Appendix A.

\begin{table*}
    \renewcommand
    \arraystretch{1.0}
    \centering
    \setlength{\tabcolsep}{8.5pt}
    \begin{tabular}{l|cccc|cccc}
    \toprule
         \multirow{2}{*}{\bf{Model}} & \multicolumn{4}{c|}{\textbf{FB15k-237 (\# 237 Relations)}} & \multicolumn{4}{c}{\textbf{WN18RR (\# 11 Relations)} }  \\ 
         \cmidrule(lr){2-5}\cmidrule(l){6-9}
          & MRR & Hits@10 & Hits@3 & Hits@1 & MRR & Hits@10 & Hits@3 & Hits@1  \\ \midrule
         TransE $[\bullet]$  & .294  & .465 & - & - &.226 &.501 &- &- \\
         DistMult $[\star]$  & .241& .419 & .263 & .155 &.43  &.49 &.44 &.39 \\
         ComplEx $[\star]$  & .247 & .428 & .275 & .158 &.44 &.51 &.46 &.41  \\
         ConvE  & .325 & .501 & .356 & .237 &.43 &.52 &.44 &.40 \\
         CapsE$[\circ]$ &.150 &.356 &- &- &.415 &.559 &- &-         \\
         RotatE   & .338  & .533 & .375 & .241 &.476  &\underline{.571} &.492 &.428  \\
         QuatE & .311  & .495 & .342 & .221 &{.481}  &{.564} &{.500} &{.436} \\ 
         DualE  &.330 &.518 &.363 &.237 &{.482} &.561 &{.500} &.440 \\
         \midrule
         R-GCN   & .248  & .417 & - & .151 &- &- &-  &-  \\ 
         SACN  & .35  & .54 & .39 & .26 &.47   &.54 &.48 &.43  \\ 
         KBGAT $[\circ]$  & .157  & .331 & - & - &.412 &.554 &- &-  \\ 
         COMPGCN  & .355 & .535 & .390 & .264 &{.479}  &.546 &{.494} &{.443}  \\ 
        RAGAT &\underline{.365} & \underline{.547} & \underline{.401} & \underline{.273} &\underline{.489}  &.562 &\underline{.503} &\textbf{.452} \\
         \midrule
        $r$-GAT &\textbf{.368} &\textbf{.558}  &\textbf{.405}  &\textbf{.276}  &\textbf{.492} & \textbf{.578} &\textbf{.506} &\underline{.449} \\ 
    \bottomrule
    \end{tabular}
    \caption{Link prediction results on FB15k-237 and WN18RR, we use RAGAT as a strong baseline. The best results are in \textbf{bold} and the second best results are in \underline{underlined}. Results of $[\bullet]$, $[\star]$ and $[\circ]$ are taken from \cite{nguyen2017novel}, \cite{dettmers2018convolutional}, and \cite{sun2019re} respectively. Other results are taken from the corresponding original papers.}
    \label{tab:lpres}
    \vspace{-4mm}
\end{table*}
\subsection{Experimental Settings}
We implement our approach with PyTorch \citep{paszke2017automatic}, and use Adam \citep{kingma2014adam} as the optimiser. The entity and relation features are learned from scratch and all the hyper-parameters are chosen by grid search based on the validation set performance. The optimal number of channels $K$ is 8 for FB15k-237, 4 for WN18RR; 2 for AIFB, 4 for MUTAG and BGS. Note that for link prediction, we fix the embedding dim for a fair comparison. Other detail settings can be found in Appendix B. We will release our source code to enhance reproducibility.

\paragraph{Evaluation Protocol} For link prediction task, we use the standard evaluation metrics: Hits@$i$ ($i \in \{1, 3, 10\}$) and mean reciprocal rank (MRR). Hits@$i$ measures the proportion of correct entities ranked in the top $i$ candidate triplets. MRR is the average of the inverse of the mean rank assigned to the correct triplet over all the candidate triplets. We adapt the \emph{filtered} setting following \citet{bordes2013translating} and use the RANDOM evaluation protocol \citep{sun2019re}.

\subsection{Baselines}
\paragraph{Link Prediction} We mainly compare our $r$-GAT with GNN-based approaches: R-GCN \citep{schlichtkrull2018modeling}, SACN \citep{shang2019end}, KBGAT \citep{nathani2019learning}, COMPGCN \citep{ vashishth2019composition}, and take a more recent approach RAGAT \citep{liu2021ragat} as a strong baseline. We also compare with several state-of-the-art KGE approaches TransE \citep{bordes2013translating}, DisMult \citep{ yang2014embedding}, ComplEx \citep{ trouillon2016complex}, ConvE \citep{dettmers2018convolutional}, CapsE \citep{vu2019capsule}, RotatE \citep{ sun2019rotate}, QuatE \citep{zhang2019quaternion}, and DualE \citep{cao2021dual}. \citet{sun2019re} find that KBGAT has a bug of test leakage during negative sampling and CapsE does not use the RANDOM evaluation protocol, so we report the results from \citep{sun2019re} where the bugs are fixed. RGHAT \citep{zhang2020relational} reports results of KBGAT with the bug before \citep{sun2019re}'s analysis, so we do not compare with its results. The results of QuatE and DualE are reported without the type constraints for a fair comparison. We do not compare with approaches using auxiliary text description here, e.g., KG-BERT \citep{yao2019kg}. 

\paragraph{Entity Classification}
We compare our $r$-GAT with Feat \citep{paulheim2012unsupervised}, WL \citep{shervashidze2011weisfeiler}, RDF2Vec \citep{ristoski2016rdf2vec}, R-GCN \citep{schlichtkrull2018modeling}, WGCN \citep{shang2019end}, RGAT \citep{busbridge2019relational} and COMPGCN \citep{vashishth2019composition}.

\begin{table}
    \renewcommand
    \arraystretch{1.0}
    \centering
    \setlength{\tabcolsep}{8pt}
    \begin{tabular}{lccc}
    \toprule
        \textbf{Model} & \tabincell{c}{\textbf{AIFB}} & \tabincell{c}{\textbf{MUTAG} } & \tabincell{c}{\textbf{BGS} }\\ \midrule
       Feat  & 55.55 & 77.94 &72.41 \\
       WL & 80.55 & 80.88 &86.20 \\
       RDF2Vec  & 88.88 & 67.20 &\underline{87.24}\\  
       R-GCN  & \underline{95.83} & 73.23 &83.10  \\ 
       WGCN  &- &77.9  &- \\ 
       RGAT &94.64 &74.15 &- \\
       COMPGCN  &- &\underline{85.3} &- \\ \midrule
       $r$-GAT (Ours) &\textbf{97.22} &\textbf{88.24} &\textbf{89.66} \\
    \bottomrule
    \end{tabular}
    \caption{Entity classification accuracy.}
\label{tab:ecres}
    \vspace{-4mm}
\end{table}

\subsection{Overall Results}
\paragraph{Link Prediction Results} From Table \ref{tab:lpres}, we find that: (1) $r$-GAT achieves state-of-the-art results on FB15k-237, and competitive results on WN18RR. The improvement indicates the efficiency of $r$-GAT by leveraging relation features to disentangle multiple semantic aspects. (2) The strong baseline RAGAT outperforms $r$-GAT's Hits@1 value on WN18RR, but their results on FB15k-237 and other metrics of WN18RR are worse than ours. We notice that FB15k-237 has more various relations and richer context than WN18RR (19 v.s. 2 edges per entity by average), which are more in line with the real-world scenarios. Therefore, the improvement on FB15K-237 indeed validates the enhanced ability of $r$-GAT to capture complex context information.

\paragraph{Entity Classification Results} As shown in Table \ref{tab:ecres}, our $r$-GAT outperforms all the previous approaches on all the datasets. This once again demonstrates the effectiveness of our disentangled approach that decouples the entity features into multiple components w.r.t the semantic aspects.

\subsection{Impact of the Different Modules}
We further conduct an ablation study on link prediction task, as shown in Table \ref{tab:dec}, we find that:
(1) Removing the query-aware attention mechanism module clearly decreases the performance, which proves the effectiveness of our proposed query-aware attention mechanism module. (2) $r$-GAT with a single-channel also decreases the performance (note that the embedding dim is fixed for a fair comparison, so each channel's dimension $\frac{D_e}{K}$ of the single-channel model is $D_e$). We suspect that the single-channel $r$-GAT degrades into a GAT-based model with relational aggregation, which only learns generic entity embeddings. At this time, the decoder cannot obtain multiple aspects' information and fails to output query-aware results. (3) Conv-TransE \citep{shang2019end} is a commonly used decoder for GNN-based approaches, here we use it to verify that combining $r$-GAT with another KGE decoder also improves the decoder's performance. Comparing with the single Conv-TransE, the results of $r$-GAT + Conv-TransE improves MRR value by a margin of 9.4\% on FB15k-237. 
(4) To further evaluate the ability of $r$-GAT to handle intertwined surrounding contexts in multi-relational graphs, we analyze the scenario of predicting missing links of entities with a high degree. This is a challenging problem because a lot of entangled information is aggregated together, and learned embeddings might suffer from an over-smoothing problem. For entities with a high degree (i.e., \textgreater 1000, which only exits in FB15k-237), $r$-GAT + Conv-TransE outperforms Conv-TransE by 61.2\% (.139 -\textgreater.224)  on Hits@10, which demonstrates the ability of $r$-GAT to handle such challenging cases.


\begin{table}
\small
    \renewcommand
    \arraystretch{1.0}
    \centering
    \setlength{\tabcolsep}{4.5pt}
    \begin{tabular}{l|cc|cc}
    \toprule
         \multirow{2}{*}{\bf{Model}}  & \multicolumn{2}{c|}{\textbf{FB15k-237 }} & \multicolumn{2}{c}{\textbf{WN18RR } }  \\ 
         \cmidrule(lr){2-3}\cmidrule(l){4-5}
          & MRR & Hits@10 & MRR & Hits@10  \\ \midrule
        $r$-GAT + Qatt & \textbf{.368} & \textbf{.558} & {.486} & {.573}  \\
        $r$-GAT (w/o Qatt) &.352 &.536 &.456 &.549 \\
        \tabincell{l}{$r$-GAT ($K=1$) + Qatt} &.349 &.533 &.457 &.558  \\
        \midrule
        $r$-GAT + Conv-TransE & {.361} & {.550} & \textbf{.492} & \textbf{.578}  \\
        Conv-TransE &.33 &.51 &.46 &.52\\
    \bottomrule
    \end{tabular}
       \caption{\label{tab:dec}Results of Ablation study.}
    \vspace{-4mm}
\end{table}

\begin{figure}
\centering  
\subfigure[FB15K-237]{
\label{Fig.sub.1}
\includegraphics[width=0.22\textwidth]{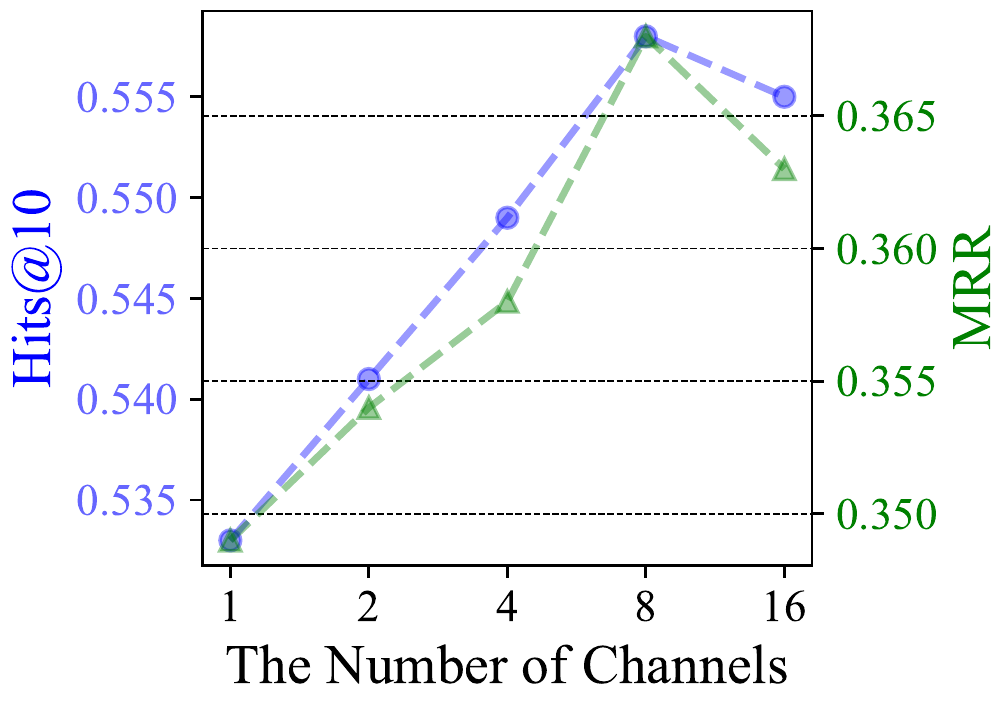}}
\subfigure[WN18RR]{
\label{Fig.sub.2}
\includegraphics[width=0.22\textwidth]{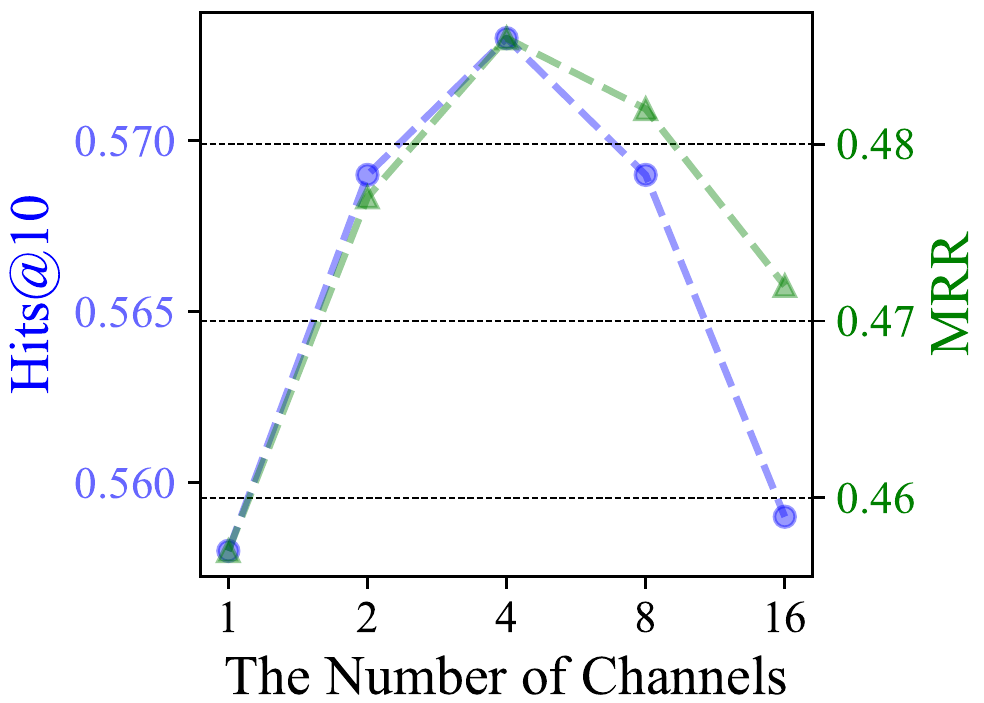}}
\caption{Impact of the number of channels.}
\label{fig:channels}
\vspace{-6mm}
\end{figure}

\subsection{Impact of the Number of Channels}
We consider varying the number of channels $K$ to run $r$-GAT with $K = $\{1, 2, 4, 8, 16\} values. From Figure \ref{fig:channels}, we find that: (1) $r$-GAT starts to perform better on both datasets when $K$ increases from 1. For example, on FB15K-237 dataset, $r$-GAT with 8 channels improves Hits@10 value by a margin of 4.7\%, and MRR value by a margin of 5.4\% comparing with the single-channel $r$-GAT (i.e., $K$ = 1). The degradation of performance validates the disadvantage of learning only generic representation. (2) However, when $K$ grows larger (e.g., $K > 8$ on FB15K-237), the performance does not increase anymore. One possible reason is the most useful information can be classified into several semantic aspects, and it is difficult to achieve a good disentanglement if the number of channels is larger than that of semantic aspects. (3) The best number of channels on WN18RR ($K$ = 4) is smaller than that of FB15k-237 ($K$ = 8). The reason is WN18RR has fewer types of relations and simpler context information, thus just need to learn a small number of channels. We select $K$ for each dataset on the validation set. Results on entity classification datasets can be found in Appendix C.

\subsection{Case Study}
In this section, we show the interpretability of our approach based on the link prediction task.

\paragraph{Relations Focus on Different Aspects.}
We choose 7 relations of FB15k-237 (more readable than WN18RR): \texttt{place\_of\_birth}, \texttt{live\_in}, \texttt{award\_nominee},  \texttt{influence\_by},
\texttt{profession},
\texttt{award\_award}, and \texttt{field\_of\_study}\footnote{Here we use short names of relations, and their full names can be found in Appendix D.} to show that results we obtain are in line with human commonsense. We randomly select 100 human entities and predict their aforementioned 7 relations. We output attention values computed by the query-aware attention mechanism (Equation \ref{eq:qa}), which represent the importance of channels to a query relation. 

\begin{table*}[ht]
    \renewcommand
    \arraystretch{1.0}
    \centering
    \setlength{\tabcolsep}{7pt}
    \begin{tabular}{l|c|c}
    \toprule
      Triplets in Test Set & $r$-GAT ($K=8$)  &  $r$-GAT ($K=1$) \\ \midrule
      \tabincell{l}{\textbf{Subject Entity}: Frank Miller \\ \\ \textbf{Query Relation}: live\_in \\ \\ \textbf{\textbf{Target}}: Los Angeles}
      &\tabincell{l}{\textbf{Top-3 Channels:}  Channel 3: 0.429,\\ Channel 5: 0.162, Channel 7: 0.115\\ \textbf{Top-4 Facts of Channel 3:} \\(Frank Miller, live\_in, Maryland) : 0.182 \\(Frank Miller, live\_in, Vermont) : 0.155\\ (Frank Miller, live\_in, New York) : 0.140\\ (Frank Miller, nationality, USA) : 0.132} 
      &\multirow{2}{*}{\tabincell{c}{\\ \\ \textbf{Top-4 Facts:} \\(Frank Miller, gender, Male) : 0.159 \\(Frank Miller, nationality, USA) : 0.124 \\(Frank Miller, marriage, Marriage) : 0.093 \\ (Frank Miller, profession, Writer) : 0.066 }}
      \\ \cmidrule(l){1-2}
      \tabincell{l}{\textbf{Subject Entity}: Frank Miller \\ \\ \textbf{Query Relation}: profession \\ \\ \textbf{Target}: Screenwriter} 
      &\tabincell{l}{\textbf{Top-3 Channels:}  Channel 6: 0.382,\\ Channel 0: 0.215, Channel 1: 0.215\\ \textbf{Top-4 Facts of Channel 6:} \\(Frank Miller, profession, Artist) : 0.200 \\(Frank Miller, profession, Writer) : 0.194 \\ (Frank Miller, profession, Author) : 0.162 \\(Frank Miller, gender, Male) : 0.154}\\ 
    \bottomrule
    \end{tabular}
    \caption{Two examples of \emph{Frank Miller}. $r$-GAT with multiple disentangled channels can let a query relation focus on more relevant known facts of an entity by focusing more on semantically related channels, thereby adapting to various query relations. These obvious contrasts indicate the importance of disentangled entity features and the query-aware attention mechanism. }
    \label{tab:frank}
    \vspace{-4mm}
\end{table*}
\begin{figure}
\centering 
\includegraphics[width=0.4\textwidth]{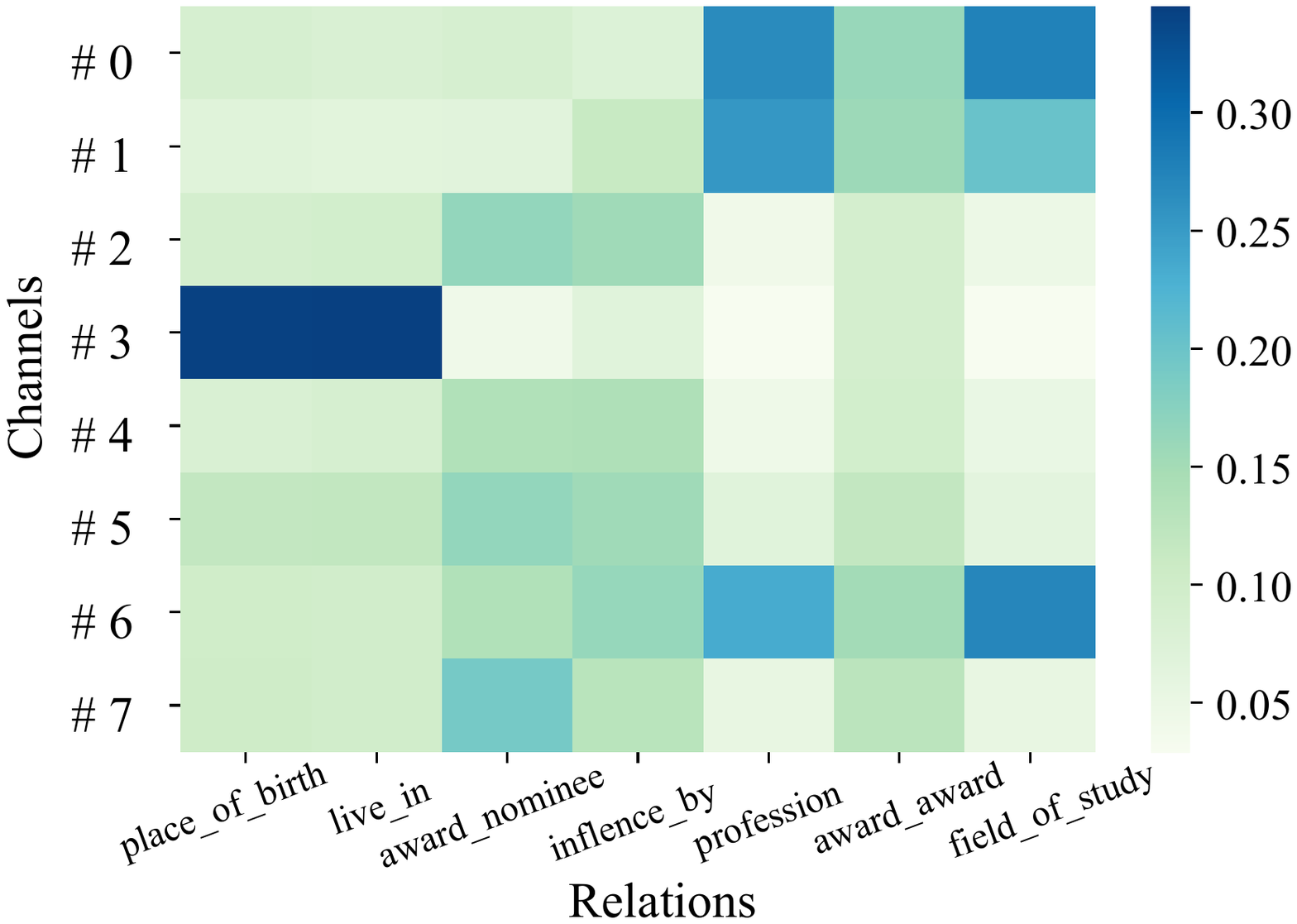}
\caption{The attention values of 8 channels to 7 relations. }
\label{fig:atts} 
\end{figure}

From Figure \ref{fig:atts}, we average the attention values of 100 entities when predicting 7 query relations. Overall, different relations focus on different channels, which matches our motivation. Interestingly, we find that the distributions of attention values are similar if the corresponding two relations are more semantically related. Specifically, \texttt{place\_of\_birth} and \texttt{live\_in} both imply places related to a person (and Channel 3 may be a disentangle aspect about "location"), so their distribution is nearly the same; \texttt{award\_nominee} and \texttt{influence\_by} both aim at people who are in the same field as a person; \texttt{award\_award}, \texttt{field\_of\_study}, and \texttt{profession} are all related to career or work that a person is engaged in, their distribution is also close. More analysis can be found in Appendix E.

\paragraph{Relations Focus on More Relevant Facts.}
Furthermore, we show that the importance of known facts to an entity is very different when predicting different relations. We take two different query relations for an entity as an example. As shown in Table \ref{tab:frank}, the known facts of \emph{Frank Miller} have different importance in the multiple disentangled channels learned by $r$-GAT. Then a query relation can focus on more relevant known facts of \emph{Frank Miller} by focusing more on semantically related channels. Therefore, the importance of facts changes dynamically with a query relation.
However, $r$-GAT with a single channel (i.e., $K = 1$) only learns generic entity embeddings for all the query relations, and the importance of the known facts is static. Therefore, it cannot output a query-aware result. Also, given that the importance of relevant facts (e.g., profession) is much smaller than the irrelevant things (e.g., gender), it is hard to explain the single-channel model's attention values.

\section{Conclusion}
In this paper, we propose a novel Relational Graph Attention network ($r$-GAT) for multi-relational graphs, which learns disentangled entity features and leverages relation features in the neighborhood aggregation. Each channel of $r$-GAT is specific to a semantic aspect of an entity. We further propose a query-aware attention mechanism to leverage different aspects' information. We prove the efficiency of different modules through extensive experiments and present the interpretability of our proposed approach. Future work will consider further exploring semantic information in relations and taking advantage of the disentangled entity features.

\clearpage
\bibliography{rgat}

\clearpage
\end{document}